\documentclass[conference]{IEEEtran}
\IEEEoverridecommandlockouts

\usepackage{cite}

\newcommand{\cred}{\textcolor{red}}

\usepackage[normalem]{ulem}
\usepackage{amsmath,amssymb,amsfonts}
\usepackage{algorithmic}
\usepackage{graphicx}
\usepackage{textcomp}
\usepackage{xcolor}
\usepackage{multirow}
\usepackage{url}
\usepackage{booktabs}
\usepackage{tikz}
\usepackage[table]{xcolor} 

\usetikzlibrary{positioning, arrows.meta}

\def\BibTeX{{\rm B\kern-.05em{\sc i\kern-.025em b}\kern-.08em
    T\kern-.1667em\lower.7ex\hbox{E}\kern-.125emX}}

\usepackage{xcolor}
\usepackage{array}
\usepackage[most]{tcolorbox}

\newtcolorbox{observationbox}{
  enhanced,
  colback=gray!10,         
  colframe=white,          
  left=6pt,                
  boxrule=0pt,             
  borderline west={2pt}{0pt}{green!80!black}, 
  sharp corners,
  fontupper=\normalsize    
}

\makeatletter
\newcommand{\linebreakand}{%
  \end{@IEEEauthorhalign}
  \hfill\mbox{}\par
  \mbox{}\hfill\begin{@IEEEauthorhalign}
}
\makeatother

\begin{document}

\title{
Instruction-Tuned LLMs for Parsing and Mining Unstructured Logs on Leadership 
HPC Systems
\thanks{This research used resources of the Oak Ridge Leadership Computing Facility at the Oak Ridge National Laboratory, which is supported by the Advanced Scientific Computing Research programs in the Office of Science of the U.S. Department of Energy under Contract No. DE-AC05-00OR22725.}
}

\author{
    \IEEEauthorblockN{Ahmad Maroof Karimi}
    \IEEEauthorblockA{
        \textit{Oak Ridge National Laboratory}\\
        Oak Ridge, USA \\
        karimiahmad@ornl.gov
    }
    \and
    \IEEEauthorblockN{Jong Youl Choi}
    \IEEEauthorblockA{
        \textit{Oak Ridge National Laboratory}\\
        Oak Ridge, USA \\
        choij@ornl.gov
    } 
    
    \linebreakand
    
    \IEEEauthorblockN{Charles Qing Cao}
    \IEEEauthorblockA{
        \textit{University of Tennessee}\\
        Knoxville, USA \\
        qcao1@utk.edu
    }
    \and
    \IEEEauthorblockN{Awais Khan}
    \IEEEauthorblockA{
        \textit{Oak Ridge National Laboratory}\\
        Oak Ridge, USA \\
        khana@ornl.gov
    }
    
}
\maketitle

\begin{abstract}
Leadership-class HPC systems generate massive volumes of heterogeneous, largely unstructured system logs. Because these logs originate from diverse software, hardware, and runtime layers, they exhibit inconsistent formats, making structure extraction and pattern discovery extremely challenging. Therefore, robust 
log parsing and mining is critical to transform this raw telemetry into actionable insights that reveal operational patterns, diagnose anomalies, and enable reliable, efficient, and scalable system analysis. Recent advances in large language models (LLMs) offer a promising new direction for automated log understanding in leadership-class HPC environments. 


To capitalize on this opportunity, we present a domain-adapted, instruction-following, LLM-driven framework that leverages chain-of-thought (CoT) reasoning to parse and structure HPC logs with high fidelity. Our approach combines domain-specific log-template data with instruction-tuned examples to fine-tune an 8B-parameter LLaMA model tailored for HPC log analysis. We develop a hybrid fine-tuning methodology that adapts a general-purpose LLM to domain-specific log data, enabling privacy-preserving, locally deployable, fast, and energy-efficient log-mining approach. We conduct experiments on a diverse set of log datasets from the LogHub repository. The evaluation confirms that our approach achieves parsing accuracy on par with significantly larger models, such as LLaMA 70B and Anthropic’s Claude. We further validate the practical utility of our fine-tuned LLM model by parsing over 600 million production logs from the Frontier supercomputer over a four-week window, uncovering critical patterns in temporal dynamics, node-level anomalies, and workload-error log correlations.

\end{abstract}


\section{Introduction}

High-Performance Computing (HPC) systems generate massive volumes of system logs, and analyzing them is critical to the sustainable and efficient operation of HPC centers~\cite{prakash2025exploration, park2017big}. 
These logs provide insights into operations at multiple levels, including application-level, system-level, hardware-level, and network-level activities. 
They help characterize the understanding of operational behavior, facilitate the diagnosis of failures, and support improvements in system reliability. However, the combination of massive data volumes and highly diverse log structures poses significant challenges to achieving accurate and efficient parsing.

The log parsing, shown as an example in Figure~\ref{fig:example_log_parsing} has long been an active area of research and engineering, with widely adopted approaches including regular expressions, rule-based systems, and structured parsing frameworks such as Drain~\cite{he2017drain} and Spell~\cite{du2016spell}. While these methods have proven effective in controlled settings, they rely heavily on heuristics and manual pattern design, making them brittle in the face of dynamic log formats, evolving templates, and domain-specific variations. The lack of labeled data and the need to continually adapt parsing logic further limit their scalability and long-term sustainability in HPC environments.

Recent advancements in log parsing have increasingly leveraged large language models (LLMs) to automate and improve the extraction of structured information from unstructured log data~\cite{prakash2025exploration}. 
Recent studies demonstrate that LLM-based methods consistently outperform traditional structured log parsers~\cite{zhong2024logparser, divlog2024}. For example, DivLog~\cite{divlog2024} uses in-context learning and LLMs to parse logs into templates, while LogParser~\cite{zhong2024logparser} combines semantic LLM insights with prefix trees for efficient clustering and template extraction. Similarly, LogAn~\cite{logan2025} deploys LLMs in production to generate structured log summaries and LogConfigLocalizer~\cite{logconfiglocalizer2024} employs a two-stage LLM strategy to localize configuration errors. 

These approaches highlight the shift toward automated, LLM-driven log parsing for improved efficiency and accuracy. However, these works use proprietary LLM models. In addition, instruction-following LLMs (a.k.a., instruct LLMs)~\cite{wei2021finetuned, ouyang2022training} and chain-of-thought (CoT) reasoning capabilities~\cite{wei2022chain} have demonstrated strong potential for tasks requiring flexible pattern recognition and contextual understanding. By leveraging these advancements, log parsing can become a more adaptive and sustainable process, capable of handling diverse log formats without constant manual intervention. Instruction-following capabilities allow LLMs to adapt parsing strategies dynamically, reducing the need for hard-coded rules and enabling more intelligent handling of unseen patterns.

One of the critical issues in applying LLMs to log parsing is data privacy. Many proprietary LLMs offer excellent performance but require users to send sensitive data to external servers for inference, which is restricted for organizations handling confidential or security-sensitive information. System administrators and HPC operators are often constrained by strict data governance policies, making it imperative to explore privacy-preserving alternatives.  Another important consideration is the economic cost of model deployment. While larger LLMs generally perform well across a wide range of tasks, their operational requirements are often prohibitively expensive. Large models such as over $100B$ parameters require several GPUs to run. 
When parsing millions of log entries, the expense of inference can become significant, making it energy inefficient and impractical. 
In contrast, for a focused task such as log parsing, a smaller or more compact LLM, when fine-tuned appropriately, can achieve performance comparable to that of a much larger model. This makes compact, fine-tuned models a more effective and practical solution, balancing accuracy with resource efficiency.

In this work, we propose a log parsing workflow built on open-source LLMs that can be fine-tuned on domain-specific log data and deployed entirely within local infrastructure. This approach eliminates the need to use proprietary LLM models and sending logs to the cloud server, ensuring compliance with privacy requirements while maintaining high parsing accuracy. We demonstrate that fine-tuning open-source LLMs for log parsing is both technically feasible and operationally effective.

In this work we propose a workflow for parsing HPC system logs powered by an instruction-following large language model (LLM). The key contributions of this work are:
\begin{enumerate}
    \item We present instruction following LLM framework which uses chain-of-thought reasoning for pattern mining from logs.  
    \item We introduce a hybrid fine-tuning methodology that combines domain-specific log templates with general instruction-following data, demonstrating that neither dataset alone achieves comparable performance.
    \item We analyze 638 million production logs from the Frontier supercomputer and present insights on failure dynamics, including cascade propagation mechanisms, topology-correlated fault localization, and science domain failure correlations.
\end{enumerate}

Note that, our approach offers privacy-preserving solution for HPC system management, avoiding the limitations of traditional methods while enabling adaptability to evolving log formats. This approach lays the groundwork for integrating LLM-powered HPC system log analysis into broader system monitoring and optimization pipelines. 


\section{Background and Related Work}

\subsection{Log Sources in HPC Systems}

\begin{figure}
    \centering
    \includegraphics[width=0.99\linewidth]{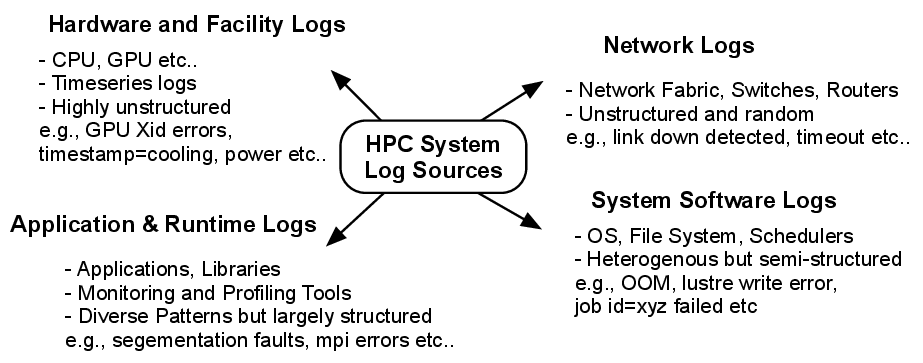}
        \vspace{-0.1in}

    \caption{Sources of logs in leadership-class HPC systems. The heterogeneity and predominance of unstructured logs especially in system monitoring with ever-changing formats motivate LLM-based parsing and pattern mining.}
    \label{fig:logsources}
\end{figure}

Supercomputer design is a multi-dimensional optimization problem that relies on extensive cross-subsystem logging~\cite{awaishpcasia, prakash2025exploration}. Figure~\ref{fig:logsources} illustrates the diverse sources of logs collected in leadership-class HPC systems such as Frontier Supercomputer~\cite{awaissc}. 
These logs originate from multiple layers, including hardware and facility components, network fabrics, system software, and application and runtime environments~\cite{park2017big}. However, note that the sources shown in the Figure~\ref{fig:logsources} are representative and do not constitute an exhaustive list. As modern HPC systems generate additional logs from monitoring services, management frameworks, and auxiliary software stacks.  Each layer produces logs with distinct characteristics. 
These range from time-series telemetry and low-level hardware errors to semi-structured system events and largely unstructured application and network messages. 
The resulting heterogeneity significantly complicates log analysis. 
This challenge is further elevated by the predominantly unstructured nature of many system logs and their continuously evolving formats, which are driven by frequent software, firmware, and hardware updates. 
Additionally, traditional log analysis pipelines struggle to cope with this variability. 
For instance, manually crafted parsing scripts and rule-based extractors are often frail and tightly coupled to specific log formats. 
They require extensive domain expertise and incur substantial maintenance overhead to accommodate new log sources and emerging system behaviors. 
Consequently, scalable and robust log parsing and pattern mining in leadership-class HPC environments require more flexible and intelligent approaches. Such approaches must be able to generalize across diverse log sources and continuously evolving system stacks.

\subsection{Traditional Log Parsing Approaches}
Log parsing aims to transform unstructured or semi-structured system logs into structured representations, enabling downstream tasks such as anomaly detection, performance monitoring, and root-cause analysis. Traditional log parsing approaches can be broadly categorized into \emph{manual rule-based methods} and \emph{automated pattern mining methods}. Rule-based techniques, such as those using regular expressions, rely on domain experts to manually define parsing rules tailored to specific log formats. While precise, these approaches require significant effort to maintain, as any change in log format demands rule updates. Automated parsers, such as Drain~\cite{he2017drain}, Spell~\cite{du2016spell}, IPLoM~\cite{makanju2009clustering}, and LogMine~\cite{hamooni2016logmine}, attempt to identify log templates using clustering or tree-based algorithms. These methods reduce manual effort but remain sensitive to noise, fail to generalize well to unseen patterns, and struggle with highly dynamic log structures. Another common approach for log parsing and analysis is to ingest log records into databases (e.g., relational, key–value, or time-series stores) and perform analysis via structured queries~\cite{park2017big}. However, database-centric approaches impose several limitations: (i) rigid schemas that struggle with evolving log formats, (ii) significant ingestion and indexing overhead for high-frequency logs, (iii) loss of contextual or causal relationships during normalization, and (iv) limited support for cross-log dependency analysis without expensive joins or custom preprocessing.

As discussed before, leadership-class HPC systems generate logs from a diverse range of sources, including job schedulers, storage subsystems, network fabrics, and compute nodes. These logs vary in verbosity, structure, and semantics, making unified parsing particularly challenging. Moreover, log formats evolve as system software and firmware are updated, and the absence of labeled datasets limits the applicability of supervised learning. 

To address these limitations, researchers have explored statistical and machine learning methods for automated log parsing. Approaches such as SHISO~\cite{mizutani2013incremental} and MoLFI~\cite{messaoudi2018search} leverage probabilistic modeling or search-based optimization to infer log structures. While these techniques improve adaptability over static rules, they still require feature engineering and often cannot capture deeper semantic relationships within logs. More recently, embedding-based representations and shallow neural models have been applied to log parsing and anomaly detection~\cite{guo2021logbert,du2017deeplog}, showing promise in template generalization but often lacking interpretability.

\begin{figure}[t]
    \centering
    \includegraphics[width=0.99\linewidth]{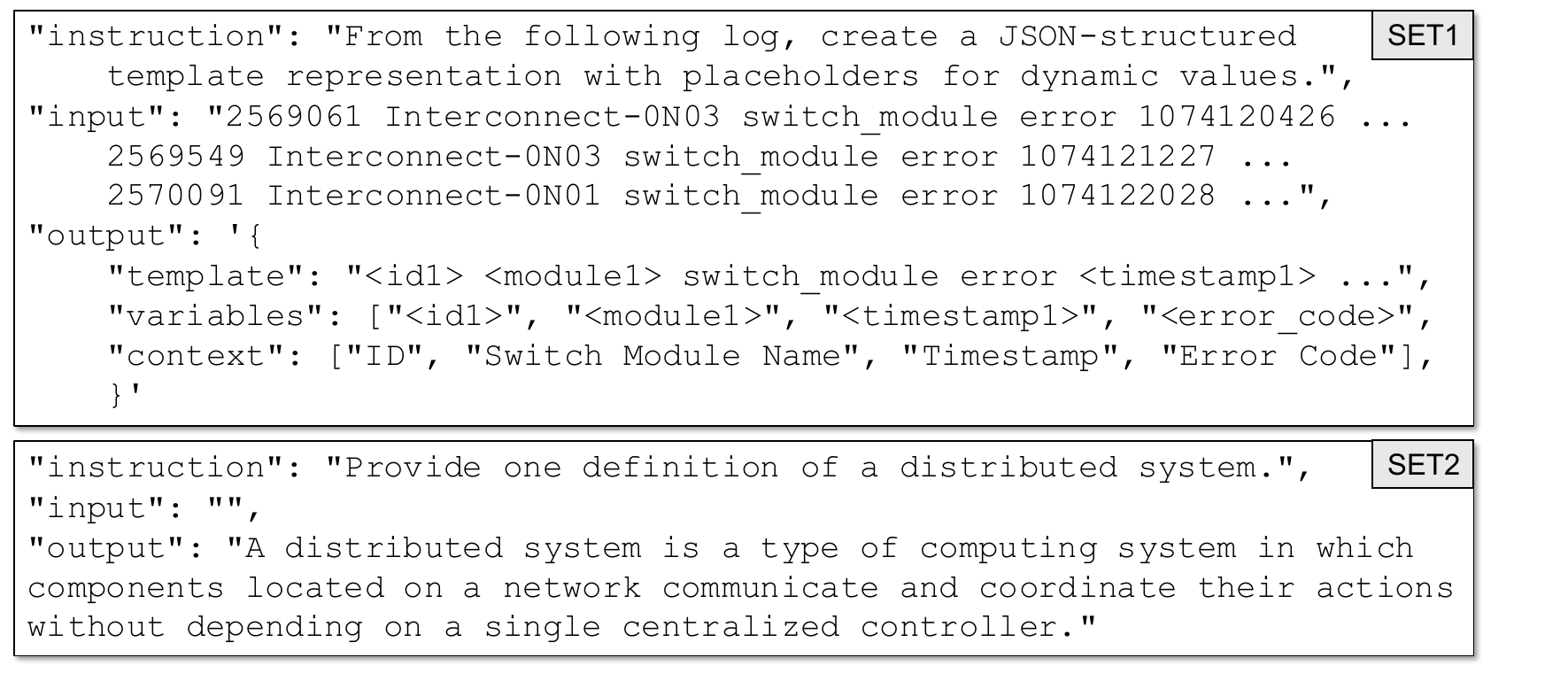}
        \vspace{-0.2in}

    \caption{Example of SET 1 and SET 2 used for fine-tuning. SET1 is generated using LLaMA 70B. SET2 is from the open-source Alpaca dataset~\cite{taori2023alpaca}. SET3 is a mixture of both.}
    \label{fig:set}
\end{figure}

\subsection{Large Language Models for Text Processing}
The LLMs GPT~\cite{brown2020language}, LLaMA~\cite{touvron2023llama}, and Falcon~\cite{almazrouei2023falcon} 
have demonstrated remarkable capabilities in zero-shot and few-shot reasoning across a wide range of natural language processing tasks. 
This LLM's capability to generalize from limited examples makes them well-suited for tasks involving unstructured text and log data. On contrary, Instruction-Following LLMs~\cite{wei2021finetuned, ouyang2022training}, fine-tuned to understand and execute human-written directives, have been particularly effective in structured information extraction. Chain-of-thought (CoT) prompting~\cite{wei2022chain} further enhances reasoning by allowing models to explicitly generate intermediate reasoning steps before producing the final output. These capabilities position LLMs as a natural fit for log parsing, where understanding both structural patterns and semantic context is critical.

Although LLMs have been widely adopted in domains such as code generation, data transformation, and document classification, their application to system log parsing is still emerging. Initial studies~\cite{ma2024llmparser,zhong2024logparser} have demonstrated that general-purpose LLMs can identify templates and extract variables from logs without extensive training, but challenges remain in adapting them for domain-specific terminology, ensuring data privacy, and controlling inference costs. This has led to growing interest in open-source LLM fine-tuning for specialized log analysis tasks. 

\section{Frontier Logs Collection and Preprocessing}

\begin{figure}[t]
    \centering
    \includegraphics[width=0.9\linewidth]{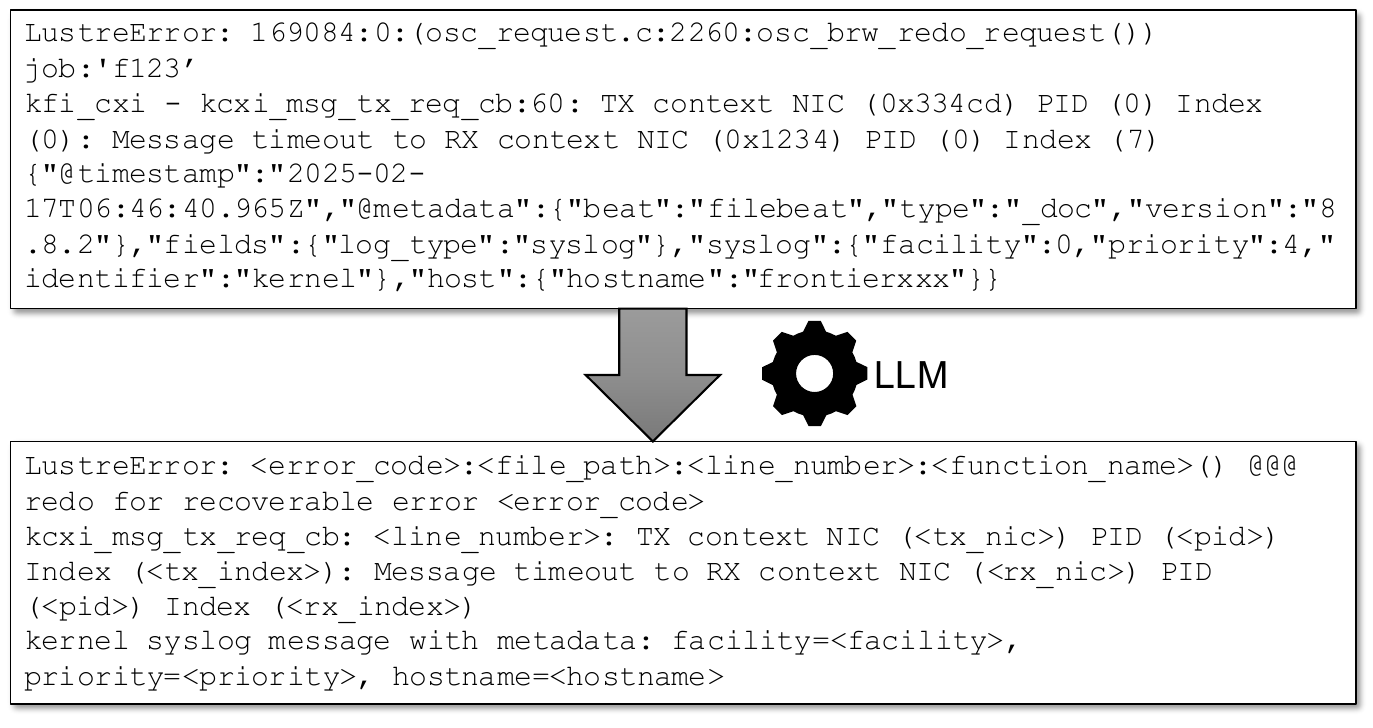}
        \vspace{-0.2in}
    \caption{An illustration of log parsing task. Given a set of raw log lines, the objective is to extract their underlying patterns, known as log templates. We propose leveraging an instruction-following LLM to achieve this.}
    \label{fig:example_log_parsing}
\end{figure}

In this section, we describe our approach to collecting and processing system logs from the Frontier supercomputer. We then preprocess these unstructured logs to extract structured patterns, enabling efficient analysis of both common and rare events. 

\subsection{System Log Collections}

We collected system logs from Frontier, an HPE Cray EX exascale supercomputer, using the HPE Performance Cluster Manager (HPCM) centralized logging infrastructure~\cite{atchley2023frontier}. The centralized logging infrastructure combines syslog messages from more than 9,400 compute nodes, storage servers, network fabric, and management infrastructure into a unified corpus. Our analysis covers the period of about 4 weeks in January to February, 2025. Over a four-week window, we collected approximately 600 million production system logs from the Frontier supercomputer. 
Raw log messages include node identifiers, timestamps, memory addresses, and unstructured text, complicating the extraction of underlying structural patterns.

\subsection{Log Preprocessing and Structural Extraction}

We employed a finetuned LLaMA 8B model to transform these unstructured logs into structured fingerprints. We show an illustrative example of this transformation in Figure~\ref{fig:example_log_parsing}.
This approach normalizes variable parameters into placeholders, enabling aggregation of semantically identical events regardless of specific node or value differences. The extracted log\_pattern serves as a fingerprint for classification and event analysis. Critically, this fingerprinting reduces millions of raw log entries to a manageable set of several thousand unique patterns, whether a pattern occurs once or hundreds of millions of times, it contributes a single row in the fingerprint table. This compression enables identification of rare anomalies that would otherwise be buried in high-volume repetitive noise. For systems of exascale size, smaller fine-tuned models can perform this extraction efficiently, processing the tens of millions of log entries that are generated daily.

In order to associate system events with specific jobs and scientific domains, we linked system logs 
with \texttt{SLURM} workload manager records. 
Note that, Frontier enforces exclusive node allocation, meaning each node runs exactly one job at any given time. This policy ensures unambiguous mapping of events recorded in the system logs to jobs based on node identity and timestamp overlap. The resulting enriched dataset combines job-level metadata (allocation account, job id, start time, end time) with system-level observations (error patterns, affected components, severity), enabling correlation analysis between workload characteristics and failure modes.
\section{LLM-Based HPC Log Parsing Framework}



\begin{figure}[t]
    \centering
\includegraphics[width=0.85\linewidth]{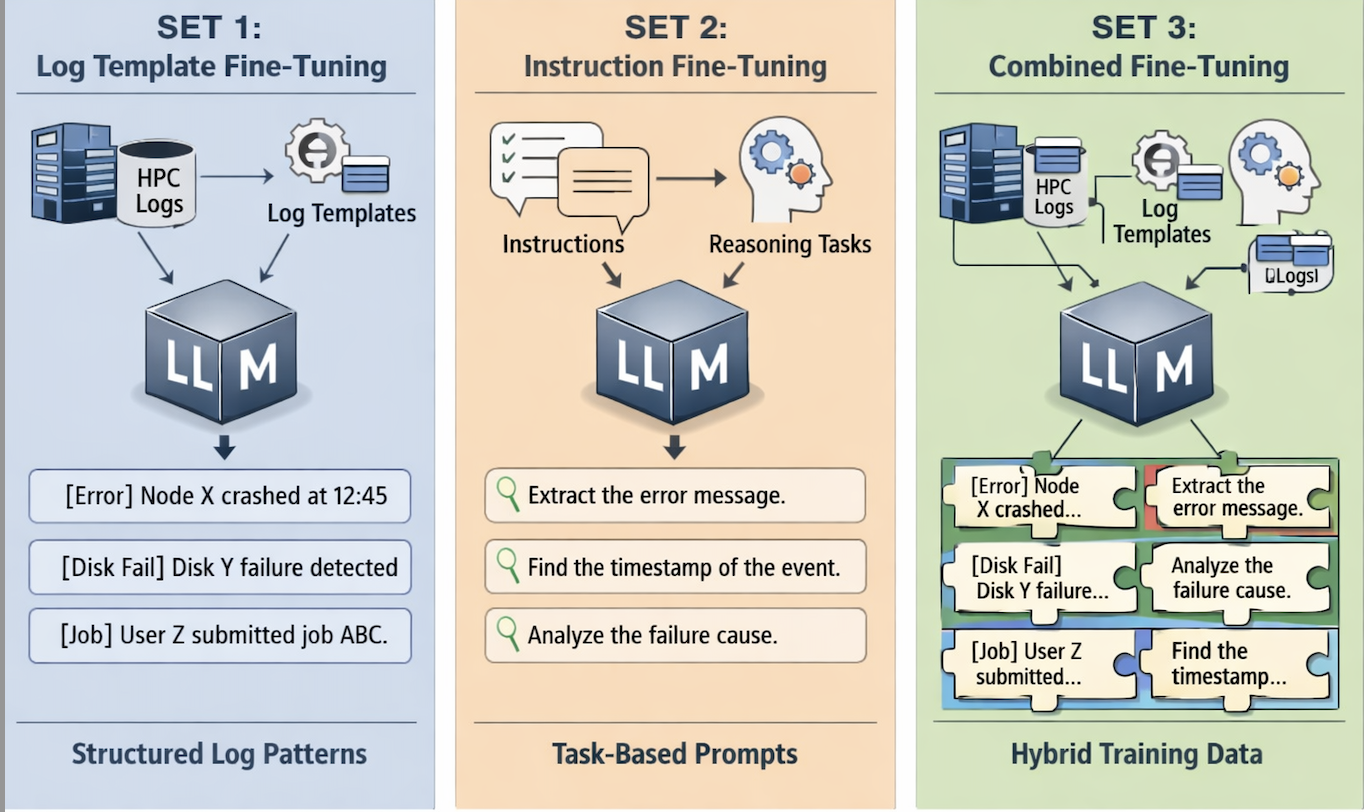}
    \caption{An overview of our proposed hybrid approach.}
    \vspace{-0.1in}
    \label{fig:overviewllmtraining}
\end{figure}

The goal of our work is to develop a scalable, accurate, and privacy-preserving framework for automated HPC log parsing and pattern mining that can generalize across diverse and evolving log sources. Therefore, we propose a framework for HPC system log parsing powered by an instruction-following LLM model. Figure~\ref{fig:overviewllmtraining} shows an overview of the proposed hybrid training methodology.
The central component of our approach is a fine-tuned, locally deployed LLM that has been adapted specifically for log template extraction and variable identification. 
The local deployment guarantees that all log data remains onsite, mitigating privacy concerns and avoiding dependence on external or proprietary LLM services.

The finetuned LLM model helps us to:
\begin{enumerate}
    \item Automatically generate log templates from raw log messages.
    \item Parse incoming logs into structured formats using the generated templates.
\end{enumerate}

This integration of a fine-tuned LLM for an end-to-end log parsing pipeline minimizes manual intervention while retaining adaptability to previously unseen log formats, enabling scalable processing of millions of HPC log entries.

\subsection{Fine-Tuning Strategy}

Our fine-tuning strategy begins with an open-source, instruction-following LLM as the base model. This choice offers several advantages. First, instruction-following models are inherently well suited for flexible log parsing tasks driven by human-readable prompts. Second, the open-source nature of the model enables full control over deployment, customization, and reproducibility. Third, fine-tuning a relatively small, task-specific model improves inference efficiency. 



We then fine-tune the base LLM model using a custom dataset. In general, two approaches can be used:
(i) leveraging domain-specific data and formats that contain paired inputs and outputs to train the LLM (e.g., for coding tasks~\cite{jain2023llm}), and
(ii) using domain-specific instructions to improve the LLM’s natural language understanding capabilities.

Accordingly, we prepare two separate datasets corresponding to each approach:
\label{sec:sets}
\begin{itemize}
    \item \textit{Log Template Dataset (SET 1):} A curated collection of log entries paired with their corresponding templates and variable placeholders. This dataset enables the model to learn the mapping from raw logs to generalized templates. 
    \item \textit{Instruction Dataset (SET 2):} A dataset designed to strengthen the model's ability to follow structured instructions, including natural language descriptions of the parsing process and desired output. This dataset is intended to improve the model’s understanding  and execution of instructions. 
\end{itemize}

\begin{figure}[t]
    \centering
    \includegraphics[width=0.7\linewidth]{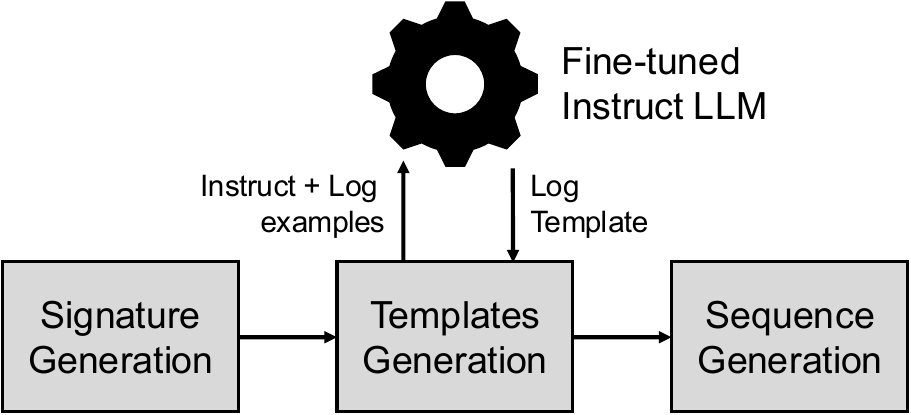}
    \caption{Inference workflow of the proposed instruction-following LLM-driven framework. Logs are processed through signature generation, template generation, and sequence generation steps, with the fine-tuned LLM employed for template generation.}
    \label{fig:inf-workflow}
\end{figure}

One key observation, as discussed in Section~\ref{subsec:sets}, is that fine-tuning the model using only SET1 or SET2 yields limited performance gains. In both cases, the resulting models exhibit only marginal improvements over the base model, indicating that each dataset in isolation lacks sufficient informational diversity to support robust log parsing.

On the other hand, jointly fine-tuning on a mixed dataset that combines \textit{SET 1} and \textit{SET 2}, referred to as \textit{SET 3}, results in a substantial improvement in parsing accuracy. This gain highlights the complementary roles of the two datasets: \textit{SET 1} provides domain-specific knowledge of log structure and variable abstraction, while \textit{SET 2} strengthens instruction-following behavior. This combination enables the model to generalize more effectively across heterogeneous log formats and parsing scenarios.




Note that, the preparation of \textit{SET 1} and \textit{SET 2} can be carried out through either manual annotation by human experts or automated data synthesis using a larger, more capable LLM, as proposed in Alpaca~\cite{taori2023alpaca}. We show the examples in Figure~\ref{fig:set}.
            \vspace{-0.1in}


\subsection{LoRA-Based Parameter-Efficient Fine-Tuning}
To perform fine-tuning efficiently, we apply the Low-Rank Adaptation (LoRA) method~\cite{hu2022lora}, a parameter-efficient fine-tuning (PEFT) technique for large language models. LoRA reduces the computational and storage cost of fine-tuning by freezing the original model parameters and injecting trainable low-rank matrices into specific weight layers (e.g., attention projection layers). During training, only these low-rank matrices are updated, while the base model remains unchanged.

Formally, given a pretrained weight matrix $W_0 \in \mathbb{R}^{d \times k}$, LoRA represents the update as:
\begin{equation}
\scriptsize
    W' = W_0 + \Delta W, \quad \Delta W = \alpha \cdot \frac{A \cdot B}{r}
\end{equation}
where $A \in \mathbb{R}^{d \times r}$ and $B \in \mathbb{R}^{r \times k}$ are trainable matrices, $r \ll \min(d, k)$ is the low rank, and $\alpha$ is a scaling factor (hyperparameter) that controls the magnitude of the low-rank update. The division by $r$ normalizes the update with respect to the rank to prevent scale explosion, while $\alpha$ allows tuning of the adaptation strength. 

\begin{figure}[t]
    \centering
    \includegraphics[width=0.99\linewidth]{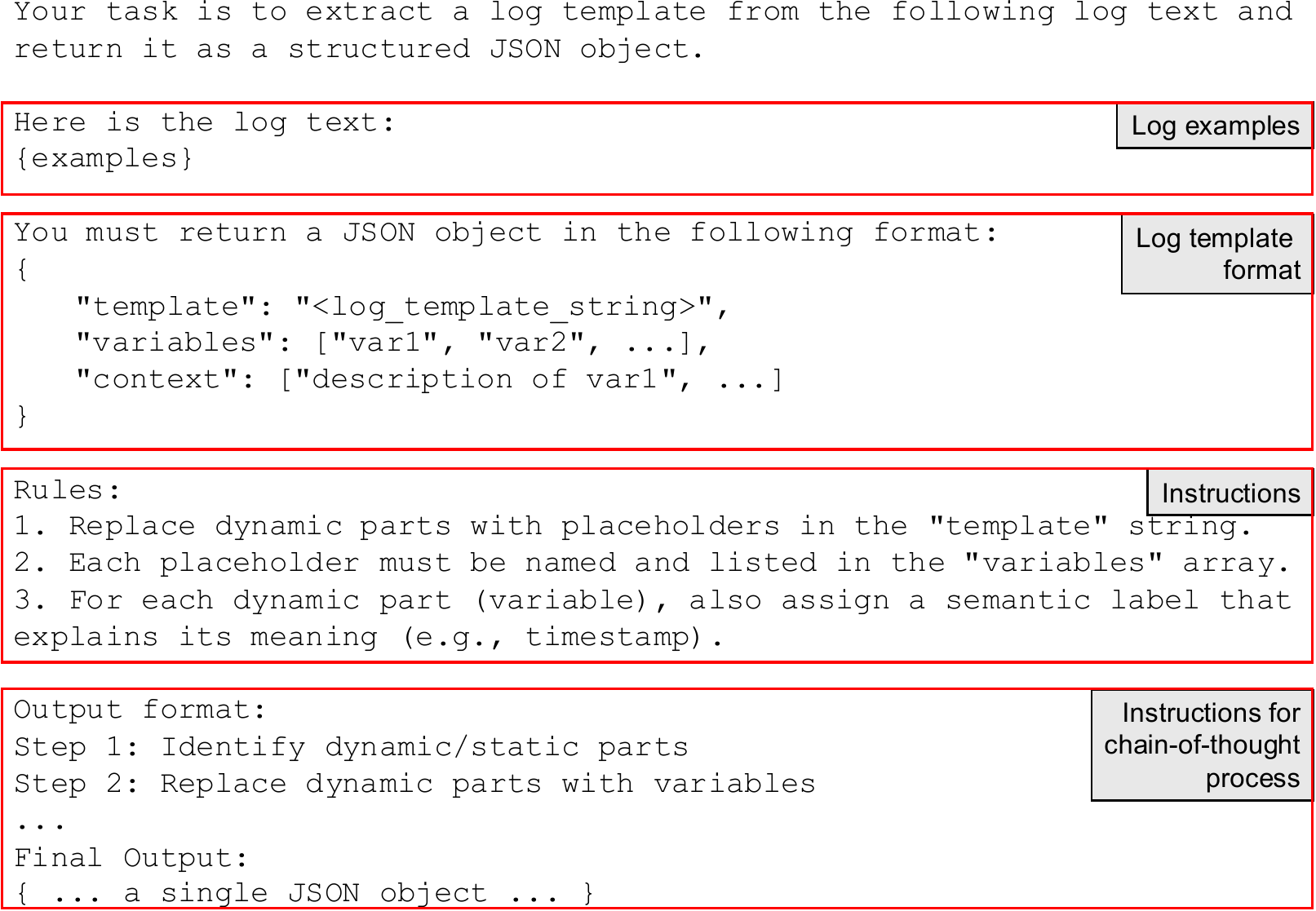}
    \caption{Simplified example prompt with instructions for log template construction and a request for the chain-of-thought process.}
    \label{fig:prompt}
\end{figure}

Selecting the appropriate rank $r$ and scaling factor $\alpha$ in LoRA is an important hyperparameter tuning decision involving multiple trade-offs. In general, a smaller rank $r$ reduces computational cost and memory usage, while a larger rank increases the model’s adaptation capacity at the expense of efficiency. Similarly, a smaller scaling factor $\alpha$ yields weaker adaptation and may underfit the task, whereas a larger $\alpha$ amplifies the influence of the LoRA weights but increases the risk of overfitting. A balanced choice of $r$ and $\alpha$ is therefore essential to achieve strong performance while maintaining generalization.

\subsection{Inference Workflow}
Once the fine-tuned model is prepared, it is integrated into a log parsing pipeline that operates in three sequential stages:

\begin{enumerate}
    \item \textit{Signature Generation:} A pre-processing stage that groups incoming logs into clusters of similar patterns based on their raw log messages. We assign each group an identifiable signature. Within each group, we randomly select $N$ representative samples to serve as an input set for the template generation.
    \item \textit{Template Generation:} For each group, we construct a structured prompt and feed it to the fine-tuned LLM, which produces generalized log templates with annotated variable placeholders.
    \item \textit{Sequence Generation:} New log entries are matched against the generated templates to produce structured, machine-readable outputs suitable for downstream tasks such as monitoring, anomaly detection, performance monitoring, or root-cause analysis.
\end{enumerate}

The prompt construction is a critical component of this workflow, particularly in Step 2. For each parsing request to the LLM, the prompt is designed to contain:
\begin{itemize}
    \item \emph{Example Logs:} A small set of representative log messages drawn from the target system.
    \item \emph{Instructions:} Natural language directives that specify how to identify templates, extract variables, and format the output.
    \item \emph{Reasoning Directive:} An explicit instruction for the model to employ chain-of-thought reasoning, improving the accuracy and consistency of the extracted templates.
\end{itemize}

The complete workflow is illustrated in Figure~\ref{fig:inf-workflow} and a simplified example of the prompt structure is presented in Figure~\ref{fig:prompt}. By combining representative examples, explicit instructions, and reasoning guidance, the model is able to parse complex and previously unseen log formats with high fidelity.



\begin{table}
\footnotesize
  \caption{List of models.}
      \vspace{-0.1in}

  \label{tab:llm}
  \centering
  \begin{tabular}{|>{\raggedright\arraybackslash}p{4.2cm}|c|l|}
    \hline
    \textbf{Model Name} & \textbf{Parameters} & \textbf{Type} \\
    \hline \hline
    Claude 3.5 Sonnet~\cite{anthropic2024claude35sonnet} & 175B & Proprietary \\
    \hline
    LLaMA 3.3 70B Instruct~\cite{meta2024llama3} & 70B & Open-source \\
    \hline
    \rowcolor{gray!40} 
    LLaMA 3.1 8B Instruct ~\cite{meta2024llama3_8b} Our Fine-Tuned Model & 8B & Open-source \\
    \hline
  \end{tabular}
    \label{tab:llm}
\end{table}
\section{Model Evaluation and Results}


\subsection{Experimental Setup and Evaluation Methodology}

We evaluate our approach using log data from the \emph{Loghub} repository~\cite{zhu2023loghub}, a widely used public benchmark for log analysis. \emph{Loghub} includes 16 heterogeneous system log datasets from domains such as HPC, Linux, OpenSSH, and Apache HTTP servers, and is commonly used to benchmark log parsing, anomaly detection, and template generation methods. For this work, we select three representative Loghub datasets: \emph{HPC}, \emph{Linux}, and \emph{OpenSSH}. Each dataset consists of 2,000 log messages. We further demonstrate the practicality of our approach by processing 638 million production logs collected over a twenty-eight-day observation window from the Frontier supercomputer, enabling large-scale log parsing and pattern mining.  



We evaluate our method using three LLM models, as listed in Table~\ref{tab:llm}:
\begin{itemize}
    \item \textit{Claude~\cite{anthropic2024claude35sonnet}:} A proprietary LLM from Anthropic, used as a high-performance reference model (upper-bound baseline).
    \item \textit{LLaMA~3 70B~\cite{meta2024llama3}:} An open-source LLM used to generate high-quality log-template datasets for fine-tuning.
    \item \textit{LLaMA~3 8B~\cite{meta2024llama3_8b}:} The base LLM model that we fine-tune with our proposed methodology. 
\end{itemize}

To evaluate the fine-tuning performance for LLaMA 8B, we prepare three distinct fine-tuning datasets that we previously discussed in Section~\ref{sec:sets}:
\begin{itemize}
    \item \textit{Log Template Dataset (SET 1): } This dataset contains 86 log messages paired with their corresponding templates. The templates were primarily generated using LLaMA~3 70B and refined through manual inspection to ensure accuracy and consistency.
    \item \textit{Instruction Dataset (SET 2):} This dataset consists of 1,000 natural language instructions paired with their expected outputs, sampled from the open-source Alpaca instruction dataset~\cite{taori2023alpaca}. It is designed to improve the LLM’s instruction-following capability while maintaining balance with SET 1.
    \item \textit{Combined Dataset (SET 3): } This dataset merges \textit{SET 1} and \textit{SET 2}, yielding 1,086 entries that provide both domain-specific log parsing knowledge and general instruction-following capability. This combination enables more robust fine-tuning and better generalization across diverse log formats.
\end{itemize}



For evaluation metrics, we measure parsing quality using the \emph{coverage rate} (\%), defined as the percentage of log messages that can be correctly parsed using the generated templates without error. 
The coverage rate is computed as:
\begin{equation}
\footnotesize
\text{Coverage Rate (\%)} = \frac{\text{\# Successfully Parsed Logs}}{\text{\# Total Logs}} \times 100
\end{equation}
This metric measures both the completeness and accuracy of the generated templates in capturing the structure of the dataset.

\subsection{Impact of Fine-Tuning Dataset on Coverage}
\label{subsec:sets}

\begin{figure}
    \centering    
    \includegraphics[width=0.99\linewidth]{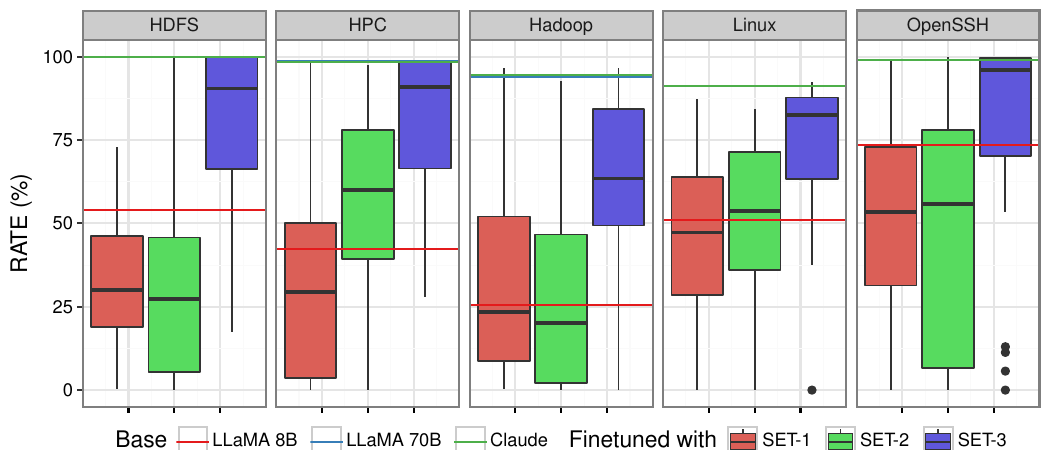}
    \caption{Overall performance after fine-tuning, compared with the three base models.}
    \label{fig:plot-mean}
\end{figure}

Figure~\ref{fig:plot-mean} presents the overall performance of the LLaMA 8B model after fine-tuning with three different datasets \textit{SET1}, \textit{SET2}, and \textit{SET3} compared against three baseline LLM models: Claude, LLaMA 70B, and the original LLaMA 8B (before fine-tuning). The evaluation is conducted on five datasets from Loghub: HDFS, HPC, Hadoop, Linux, and OpenSSH. Reported coverage rates are averaged over multiple fine-tuned models trained with different LoRA hyperparameter configurations.

For the LLaMA 8B fine-tuning experiments, we vary the number of epochs (2, 3, 5, 10), LoRA rank (8, 16), and LoRA scaling factor $\alpha$ (8, 16, 32). This variation allows us to assess the robustness of fine-tuning performance across a range of parameter-efficient training settings.

\begin{figure}
    \centering
    \includegraphics[width=0.99\linewidth]{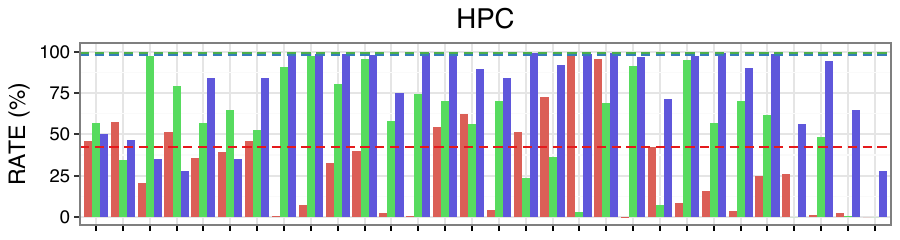}
    \includegraphics[width=0.99\linewidth]{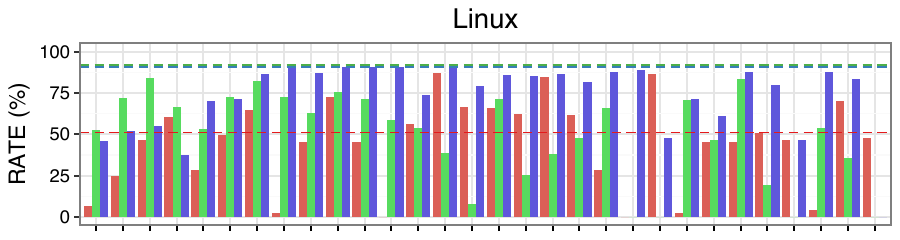}
    \includegraphics[width=0.99\linewidth]{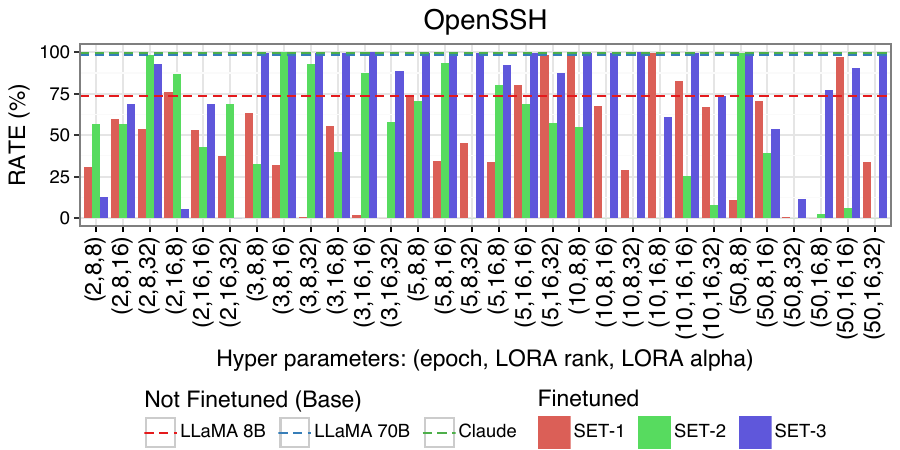}
    \caption{Full experimental results for the fine-tuned LLaMA 8B model using different training sets and varying LoRA hyperparameters, training epochs, LoRA rank, and LoRA alpha.}
    \label{fig:plot-all}
\end{figure}

Figure~\ref{fig:plot-all} presents the full experimental results for HPC, Linux, and OpenSSH logs using the fine-tuned LLaMA 8B model. As before, the LLaMA 8B model is fine-tuned with different training datasets and a range of LoRA hyperparameter configurations. The performance of base models are plotted in the dotted lines.

First, we observe that 
fine-tuning LLaMA 8B enables it to achieve performance comparable to much larger general purpose models, such as Claude and LLaMA 70B, thus making our approach energy and computationally efficient. Second, training with SET3 (the combination of SET1 and SET2) consistently yields better results in many cases, highlighting the benefit of combining domain-specific and instruction-following data. Finally, longer training schedules do not necessarily improve performance; in fact, shorter runs of 3–5 epochs often perform better, likely because they avoid overfitting and preserve useful knowledge from the base model.

\subsection{Validation on Production-Scale Frontier System Logs}

\begin{figure}
    \centering
    \includegraphics[width=0.9\linewidth]{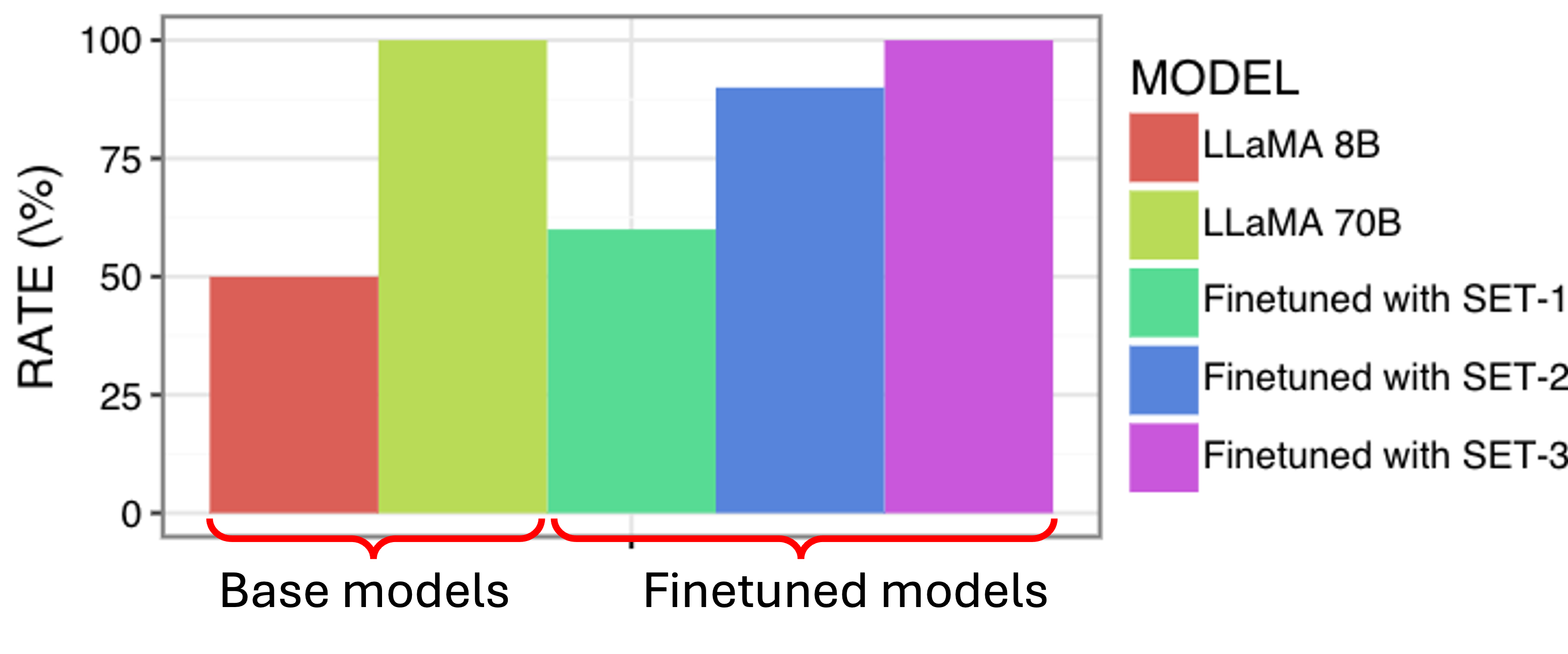}
    \caption{Evaluation on sampled Frontier system logs. The base models (left) are compared against the LLaMA-3 8B models finetuned on various sets (right). Notably, the model finetuned with SET-3 (LoRA (3, 8, 8)) achieves an accuracy rate comparable to the much larger LLaMA 70B base model.}
    \label{fig:plot-frontier}
\end{figure}

In our next experiment, we selected one of the best-performing models used above and applied it to real-world Frontier system logs. Specifically, we used a pretrained LLaMA3 8B model fine-tuned with LoRA hyperparameters (epoch = 3, LoRA rank = 8, LoRA alpha = 8) and evaluated it on a randomly selected sample of 25K Frontier logs. Figure~\ref{fig:plot-frontier} presents the results. We observe the same trends on the Frontier data: the fine-tuned LLaMA3 8B with SET3 achieves accuracy comparable to the much larger LLaMA 70B model.

\subsubsection{Fine-Tuned LLM Model Cost Comparison}

We measure the cost of running LLMs  on HPC systems to demonstrate that fine-tuned, smaller LLMs can be both practical and economical. We profile memory usage and energy consumption for inference with LLaMA 8B and LLaMA 70B on Frontier at Oak Ridge National Laboratory. Frontier nodes contain 4 GPUs, each consisting of 2 AMD MI250X GCDs with a total of 64 GB of high-bandwidth memory (HBM) per GPU.

\begin{figure}
    \centering    
    \includegraphics[width=0.9\linewidth]{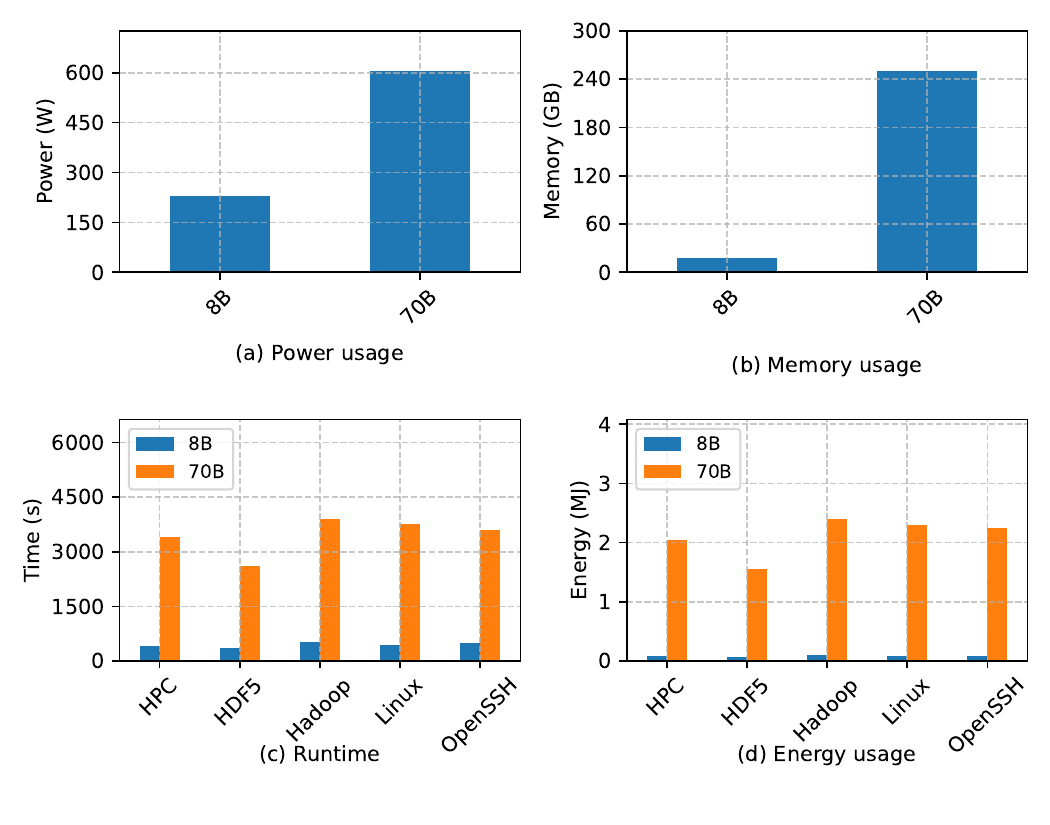}
        \vspace{-0.35in}
    \caption{Resource cost comparison of LLaMA models (8B and 70B) on Frontier supercomputer (a) average power, (b) average memory, (c) execution time, and (d) total energy consumption.
   }
    \label{fig:plot-profiling}
\end{figure}

Figure~\ref{fig:plot-profiling} presents the profiling results. On average, the LLaMA 8B model consumes 235 Watt (W) of power and 20.5 GB of memory, whereas the LLaMA 70B model consumes 609 Watt (W) and 252 GB of memory. Inference with LLaMA 8B takes 495 seconds on average, while LLaMA 70B requires about 3,493 seconds, which is more than 7x longer.
As a result, the total energy consumption for a full inference run is approximately 0.1 Megajoule (MJ) for LLaMA 8B, compared to 2.1 Megajoule (MJ) for LLaMA 70B, which is an increase of more than 20x.

\subsection{Operational Insights from Large-Scale LLM-Parsed Logs}
    \label{sec:analysis}

    \begin{figure}
        \centering
        \includegraphics[width=\linewidth]{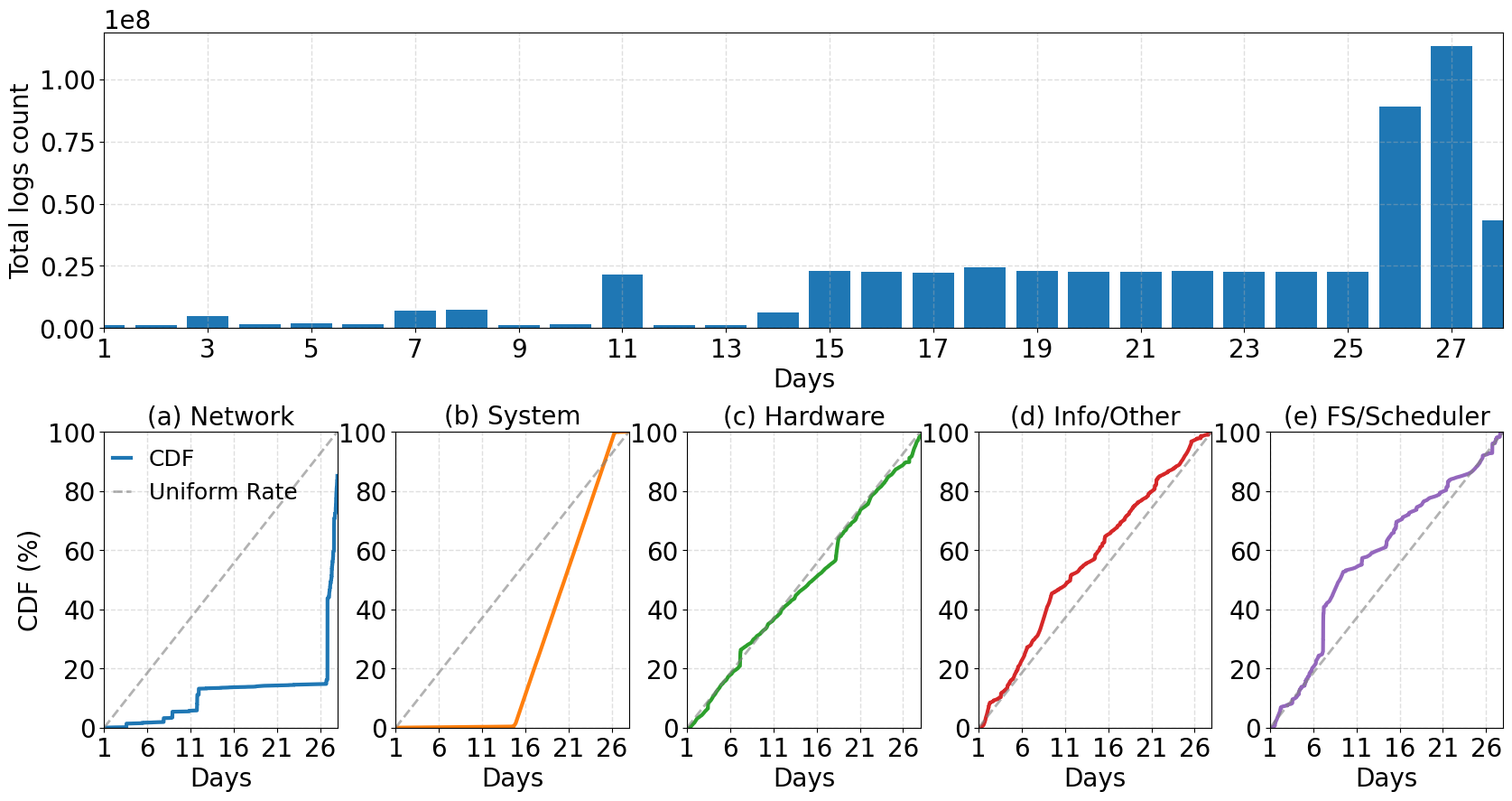}
        \caption{Temporal distribution of system logs showing overall frequency (top) and cumulative distribution by category (bottom) over a four-weeks analysis period on Frontier supercomputer.}
        \label{fig:num_logs}
    \end{figure}
    
    We applied our fine-tuned LLM to analyze $638$ million system logs from Frontier over a four-week observation period. Our analysis reveals operationally significant patterns in failure dynamics, cascade propagation, and workload-failure correlations.

    \begin{table}[t]
        \centering
        \caption{Log severity distribution on 600M+ parsed logs by LLM.}
            \vspace{-0.1in}
        \label{tab:log_levels}
        \scriptsize            \setlength{\tabcolsep}{3pt}

        \begin{tabular}{lrrl}
        \toprule
        \textbf{Category} & \textbf{Count} & \textbf{\%} & \textbf{Insight} \\ \hline
        \midrule
        INFO              & 260.1M & 40.7 & Baseline activity \\
        ERROR             & 189.8M & 29.7 & \multirow{2}{*}{50 \% are errors} \\
        DISK ERROR        & 129.6M & 20.3 & \\
        WARNING           & 39.8M  & 6.2  & Degradation signals \\
        HARDWARE ERROR    & 15.1M  & 2.4  & Silent corrections \\
        \midrule
        CRITICAL/FATAL    & 450K   & 0.07 & Visible failures \\
        KERNEL PANIC/CRASH& 2.3K   & 0.0003 & Unrecoverable \\
        \bottomrule
    \end{tabular}
    \vspace{1mm}
    
    \end{table}
    
    \subsubsection{Log Severity Distribution and Cascading Effect}

        Table~\ref{tab:log_levels} presents the severity distribution of log entries. A striking finding is realizing the \textit{needle in the hay stack}. The catastrophic events such as {\footnotesize KERNEL PANIC, SYSTEM CRASH} represent only $0.0004\%$ of logs, they are hidden by $15.1$ million silent hardware corrections. This huge imbalance quantifies the hidden reliability burden that for every unrecoverable system crash log, approximately 6,500 recoverable hardware events ({\footnotesize ECC} memory corrections, PCIe retries) accumulate in the background on top of other errors and info level log. \textit{Our observation from this analysis is that this substantial window between warning signals and catastrophic failure presents a concrete opportunity for predictive maintenance interventions.}


        
        The frequency of error-level messages ({\footnotesize ERROR + DISK ERROR}) account for 50\% of all logs, outnumbering informational messages (40.7\%). This distribution indicates that Frontier's logging infrastructure is tuned for failure detection rather than routine status reporting, a design choice that facilitates the pattern mining approach presented in this work.
    
    \subsubsection{Temporal Dynamics and Failure Periodicity}
        Figure~\ref{fig:num_logs} reveals two operationally significant temporal patterns. First, log volume exhibits \textit{bursty behavior}, fluctuating by orders of magnitude between consecutive 5-minute windows. These spikes correlate with incident events rather than gradual degradation, suggesting that failures on Frontier tend to manifest abruptly rather than through slow accumulation.
        
        Second, the cumulative distribution function (CDF) shows that different type of log exhibit different frequency characteristic. For example network log shows that about 90\% of the logs were generated in 1-2 days also contributing to a several order of fluctuation in the temporal plot at the top. The CDFs of Figure~\ref{fig:num_logs} (c)-(e) show a relatively uniform progression showing gradual degradation eventually leading to severe degraded state. \textit{Our analysis from this is that Compounding instability: once the system enters a degraded state, error rates accelerate rather than stabilize, underscoring the importance of early intervention before failure cascades take hold.}


        
    \subsubsection{Anatomy of a Cascade Failure}

        \begin{figure}
            \centering
            \includegraphics[width=0.99\linewidth]{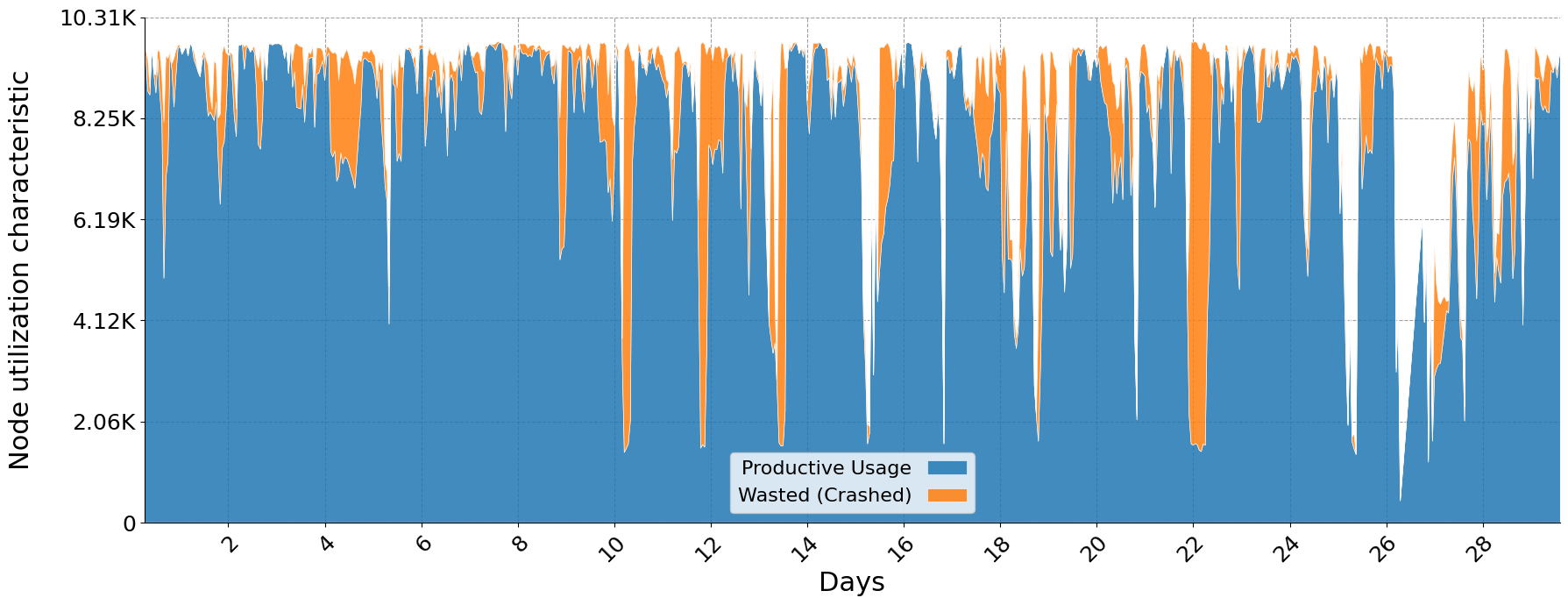}
                        \vspace{-0.1in}
            \caption{
            \footnotesize
            The distribution of Frontier compute node-hour over the four week period. The proportion of blue color shows the percentage of node utilization by successfully completed jobs(83.17\%), the orange color shows the wasted node utilization by failed jobs(9.32\%). The white space shows unutilized compute nodes in terms of node-hours(7.53\%). }
            \label{fig:motivation}
        \end{figure}
        
        \begin{figure}
            \centering
            \includegraphics[width=0.95\linewidth]{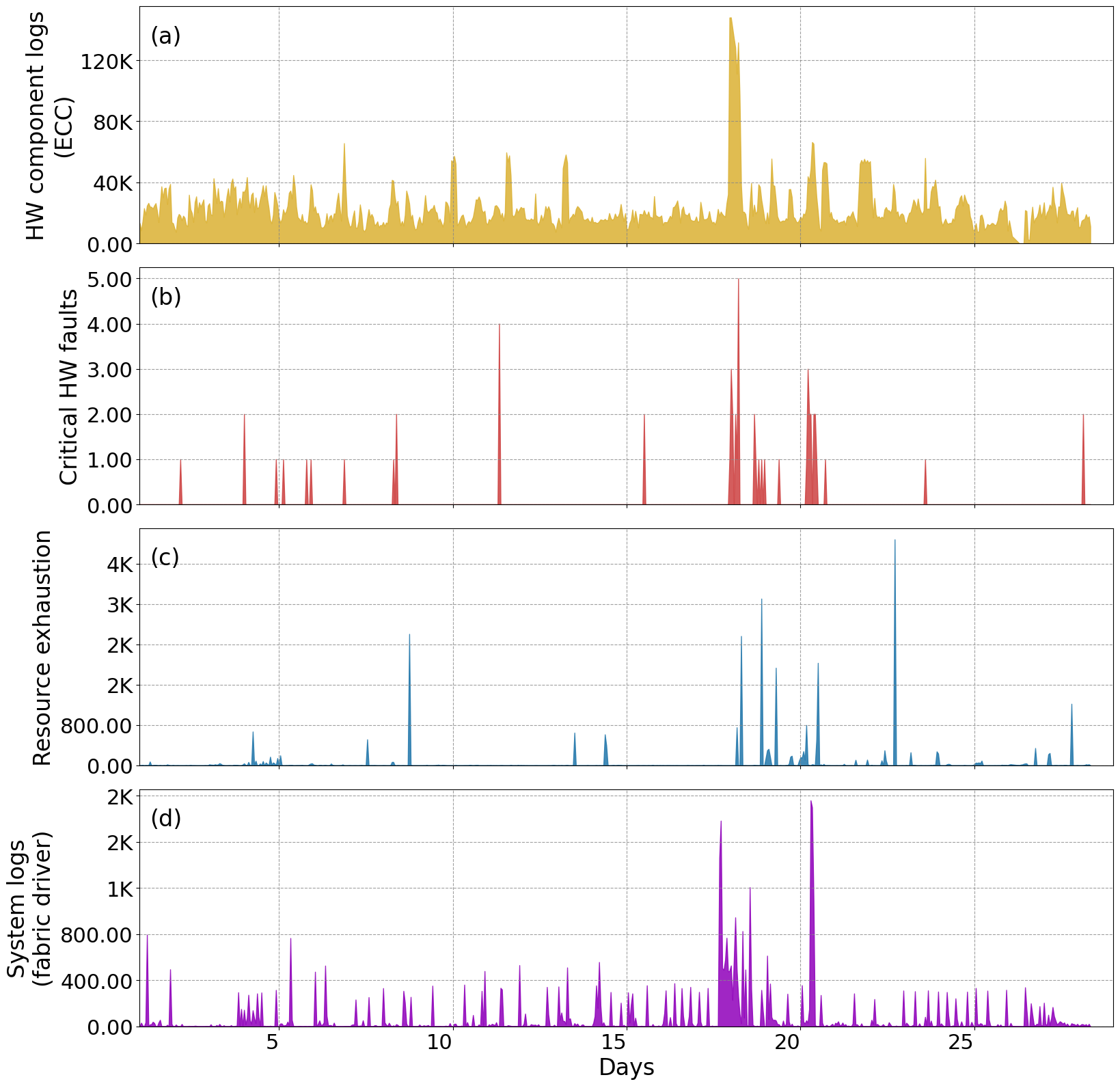}
                \vspace{-0.1in}

            \caption{
            \footnotesize
            System logs showing cascading events. The timeline illustrates the propagation of failure  the physical layer to the system layer. Hardware (HW) component events (a) (15.1 million ECC/retry events) indicates hardware stress. A spike in Hardware Faults (plot (b)) often initiates a system-wide cascade, resulting in concurrent spikes across all other monitored metrics. This loss stalls I/O operations, causing Resource Exhaustion (c) as pending data saturates system memory, triggering over 100,000 OOM errors. Plot(d) shows system level error.}
            
            \label{fig:temporal_corelation_of_events}
        \end{figure}
        One the significant finding from our analysis is the identification of a four-stage failure cascade that explains how isolated hardware faults escalate into significant outage captured by deep white valleys on day 18-19 and few others on day 10, 12, and 14 as shown in Figure~\ref{fig:motivation}.
        
        \textit{Silent Degradation:} The cascade begins with a chronic baseline of 15.1 million recoverable hardware events, primarily ECC memory corrections and PCIe bus retries. These events are silently corrected and do not immediately impact running jobs, but they signal accumulating hardware stress, see Figure~\ref{fig:temporal_corelation_of_events}-(a).
        \textit{Critical Trigger:} A small number of critical hardware faults (55 events in our observation window) mark the transition from degraded to failing, see Figure~\ref{fig:temporal_corelation_of_events}-(b). These include double-bit memory errors and fatal bus errors that cause definitive loss of components such as GPUs or storage controllers marked by spikes in Figure~\ref{fig:temporal_corelation_of_events}-(a).
        \textit{Resource Exhaustion:} In Figure~\ref{fig:temporal_corelation_of_events}-(c), as failed components stall I/O operations, pending data accumulates in system memory. This triggers massive memory pressure, forcing the kernel's OOM killer to terminate over many processes, a significant spike in error logs is observed in Figure~\ref{fig:temporal_corelation_of_events}-(a).
        \textit{Fabric Collapse:} The cascade culminates in system-level failures, with over 90\% caused by deadlocks in the \textit{kkfilnd} fabric driver. Detailed log pattern analysis confirms recurring stack traces pointing to the Kernel \textit{KoNET} Fabric Interface \textit{LNet} driver as the terminal failure point. \textit{Our observation is that this multi-stage progression demonstrates that a single critical component failures (only 55 recorded instances) in tightly-coupled HPC systems propagate into widespread fabric deadlocks, a finding enabled by the semantic pattern extraction capabilities of our fine-tuned LLM.}


    \subsubsection{Workload-Failure Correlation}
\begin{figure}
            \centering
            \includegraphics[width=0.95\linewidth]{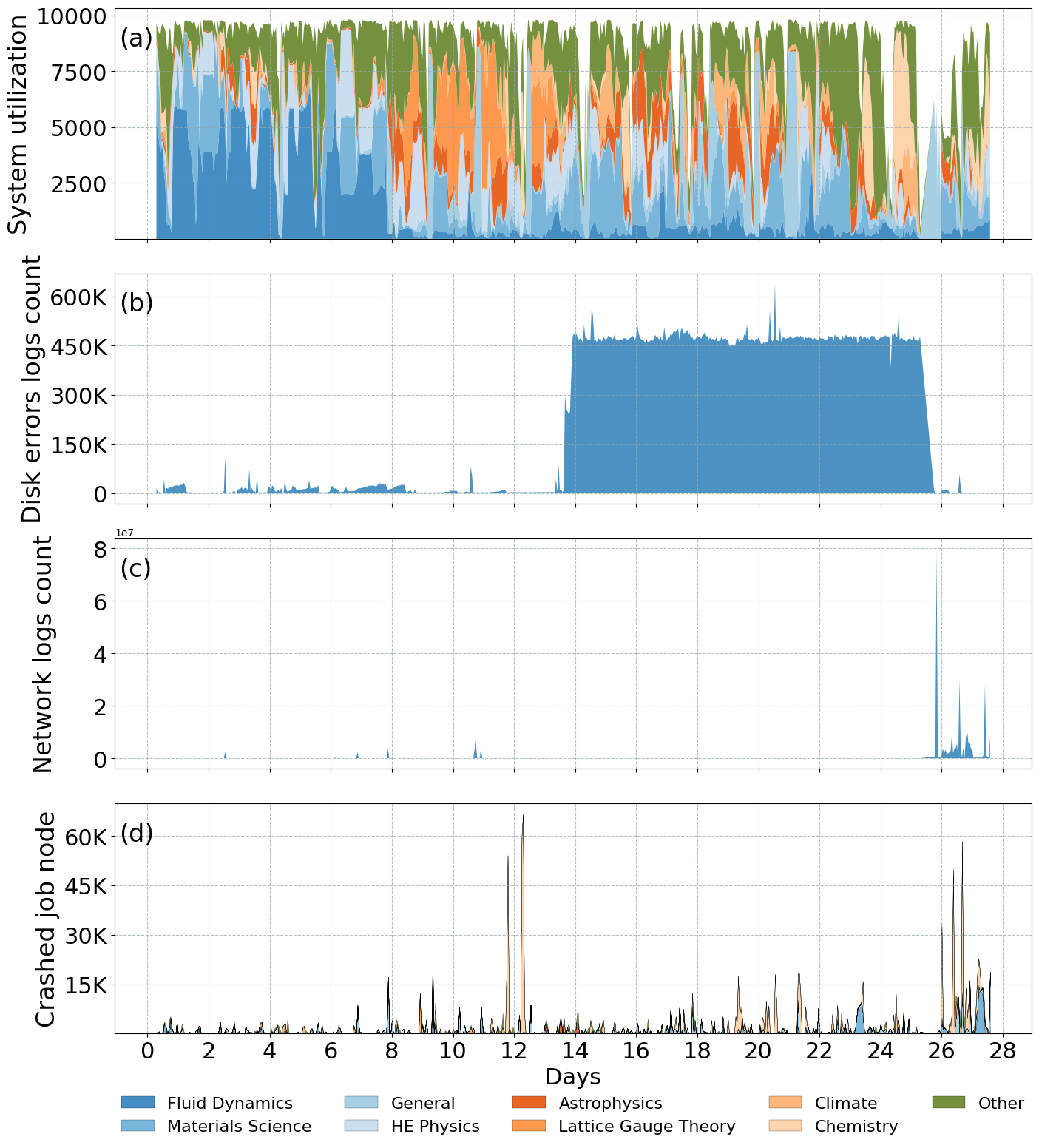}
                \vspace{-0.1in}

            \caption{Correlation of scientific domain workload with system instability events captured by the system logs. Jobs with most frequent science domain are shown with different colors. }
            \label{fig:network_corelation}
        \end{figure}

        Figure~\ref{fig:network_corelation} correlates scientific domain workload with system instability events. Plot~(a) shows compute node utilization stratified by scientific domain, providing context for interpreting the error patterns in subsequent plots(b)--(d). 
        
        \textit{Decoupling of Error Volume from Job Failures:} 
        A striking finding emerges from plots Figure~\ref{fig:network_corelation} (b)--(d). Despite 400,000 disk errors per 5-minute window plot (b), which is equivalent to over 115 million disk errors logs daily, job failures (plot (d)) and network disruptions (plot (c)) remained minimal. 
        Log pattern analysis identifies these as \texttt{rsyslog} write failures, the system logging daemon on compute nodes encountered this local storage exhaustion. However, the HPCM infrastructure continued receiving system events via network transport. This finding illustrates that error volume alone is insufficient for predicting operational impact, the affected subsystem determines whether failures remain contained or propagate.
    
        \textit{Slingshot Failure Spike:} The network spike on day 26 captured in plot (c) (reaching $8 \times 10^7$ events) demonstrates the contrast. Unlike logging errors, fabric disruptions propagate immediately to running jobs because MPI-synchronized workloads depend directly on interconnect availability. The correlated spike in job terminations across scientific domains (plot (d)) confirms that the Slingshot interconnect constitutes a single point of failure for MPI-collective operations.
        \textit{Operational stability is determined more by the resilience of the affected subsystem than even by the massive volume, persistent disk logging errors can remain contained, but even brief interconnect disruptions cause immediate job terminations.}

    \subsubsection{Failure Localization and Network Topology}
\begin{figure}
            \centering
            \includegraphics[width=0.95\linewidth]{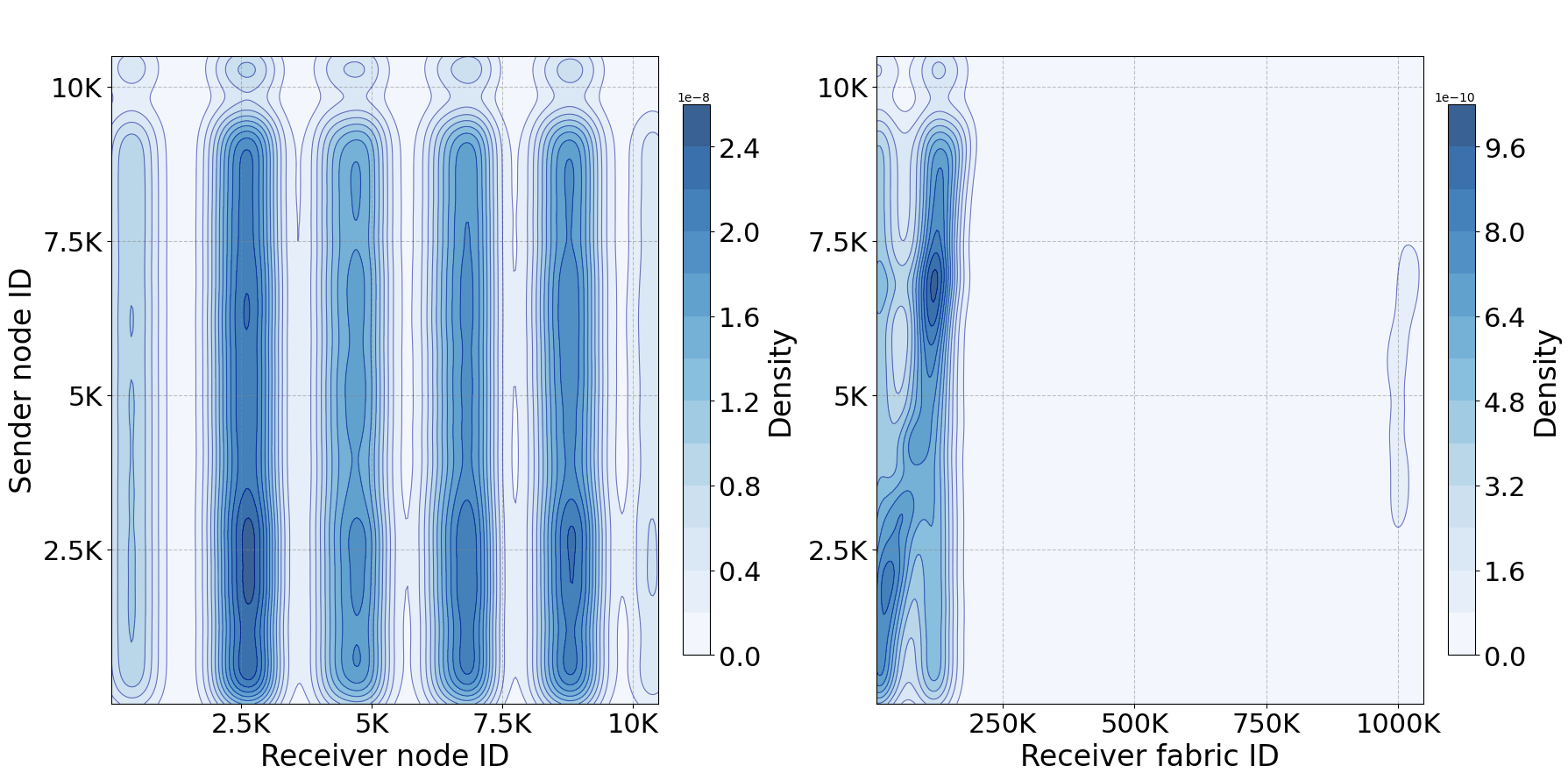}
            \vspace{-0.1in}

            \caption{ 
                Kernel density estimation~\cite{chen2017tutorial} of interconnect error distribution across Frontier's network address space. Color intensity indicates error concentration between sender-receiver pairs. \textit{(Left)} Node-to-node communication fault. \textit{(Right)} Node-to-fabric errors.}
                \label{fig:spatialAnalysis}
        \end{figure}

        Figure~\ref{fig:spatialAnalysis} maps interconnect errors across Frontier's network address space using kernel density plot~\cite{chen2017tutorial}, revealing the spatial structure of failure propagation.  Color intensity indicates error concentration between sender-receiver pairs. The left-plot shows the node-to-node communication error and the right-plot describes node-to-fabric communication error. 
        
        \textit{Node-to-Node Communication:} The error density in left plot exhibits vertical banding at about 2,000 to 2500 node intervals, corresponding to Frontier's hierarchical network topology where nodes are grouped by rack and cabinet. This vertical structure indicates \textit{receiver-localized} failures, when specific node ID ranges become unreachable, all active senders across the system (Y-axis spanning 0 to 10K) simultaneously log errors against those targets. The uniform vertical extent confirms that these outages manifest as global events, a single infrastructure failure (e.g., Rows of Rack, Top-of-Rack switch) renders an entire node group unreachable to the full system rather than causing isolated pairwise communication failures.
        
        \textit{Node-to-Fabric Communication:} Errors targeting fabric infrastructure (switches, routers with IDs exceeding the compute node range) exhibit markedly different characteristics. The error density shows that failures concentrate at low fabric IDs with an isolated cluster near fabric ID 1M. This distribution 
        indicates that failures predominantly affected \textit{edge infrastructure} rather than the core fabric 
        backbone. The sparsity at mid-range fabric IDs (250K--750K) confirms that Frontier's Slingshot core remained operationally stable throughout the observation period. \textit{Our observations reveals that the edge-dominated failure pattern suggests that reliability investments targeting network edge components (Top-of-Rack switches, node injection bandwidth) may yield greater operational improvements than core fabric hardening. Additionally, the regular periodicity of node-level failures provides a basis for topology-aware job placement scheduler policies.}

 \begin{table*}[t]
    \centering
    \caption{Performance Metrics and Pattern Extraction Failures by Perturbation Type}
    \vspace{-0.1in}
    \label{tab:llm_robust}
    \scriptsize
    \renewcommand{\arraystretch}{1.2}
    \setlength{\tabcolsep}{3pt}
    \begin{tabular}{@{}lccccccc@{}}
        \toprule
        \textbf{Type} & \textbf{Acc.} & \textbf{Avg. Sim.} & \textbf{Lev.} & \textbf{WER} & \textbf{Original Log Pattern} & \textbf{Transformed Log Pattern} & \textbf{Error} \\ \hline
        \midrule
        Param Change  & 96.5\% & 0.997 & 0.025 & 5.7\% & \texttt{"@ts":"<ts>"...:"<host\_h>"} & \texttt{"@ts":"<ts>"...:"<h>"} & \texttt{<host\_h>} $\to$ \texttt{<h>} \\
        Typo          & 94.5\% & 0.996 & 0.004 & 3.4\% & \texttt{out of memory: killed...} & \texttt{ovt of memory: killed...} & \texttt{out} $\to$ \texttt{ovt} \\
        Whitespace    & 78.8\% & 0.978 & 0.061 & 10.0\% & \texttt{<m> - <f>:<l>: tx...} & \texttt{<m>-<f>:<l>: tx...} & Space removed \\
        Word Reorder  & 77.7\% & 0.974 & 0.041 & 3.5\% & \texttt{<m>-<f>:<l>: tx nic...} & \texttt{posted=<v> <m>-<f>:<l>...} & Token prefix \\
        Punctuation   & 57.1\% & 0.864 & 0.016 & 0.0\% & \texttt{...nic (<id>) pid...} & \texttt{...nic [<id>] pid...} & () $\to$ [] \\
        Missing Words & 44.2\% & 0.962 & 0.040 & 3.9\% & \texttt{out of memory: killed...} & \texttt{of memory: killed...} & "out" missing \\
        Extra Words   & 8.5\%  & 0.844 & 0.052 & 5.2\% & \texttt{<m>-<f>:<l>: tx nic} & \texttt{warning: <m>-<f>:<l>: tx nic} & Prefix added \\
        \bottomrule
    \end{tabular}
\end{table*}
{
        
        \begin{table}[t]
            \scriptsize
            \centering
            \caption{Log error category groups used in Figure~14 Y-axis. Categories are grouped by affected subsystem.}
                \vspace{-0.1in}
            
            \label{tab:error_categories}
            \setlength{\tabcolsep}{6pt}
            \renewcommand{\arraystretch}{1.15}
            \begin{tabular}{ll|ll}
            \toprule
            \textbf{ID} & \textbf{Category} & \textbf{ID} & \textbf{Category} \\
            \midrule \hline
            \multicolumn{4}{l}{\textit{Network-related}} \\
            AA & Network: Timeout            & AP & Network: CXI Protocol \\
            AB & Network: Refused            & AU & Network: Force Close \\
            AC & Network: Connect Fail       & AX & Network: PCIe        \\
            AE & Network: CXI Hardware       & AZ & Network: Timeout     \\
            AK & Network: CXI Cache          &    &                      \\
            \midrule
            \multicolumn{4}{l}{\textit{Hardware-related}} \\
            AF & Hardware: ECC Warning       & AQ & Hardware: Component  \\
            AJ & Hardware: Driver            & AW & Hardware: Firmware   \\
            AO & GPU: VM L2 Fault            & AV & GPU: General         \\
            \midrule
            \multicolumn{4}{l}{\textit{System and Scheduler}} \\
            AD & System: Service Failure     & AH & System: Log Dropped  \\
            AL & System: DHCP                & AM & Scheduler: Daemon    \\
            AS & System: Disk Error          & BA & System: Resource     \\
            AT & System: Stack Trace         & BB & System: OOM          \\
            \midrule
            \multicolumn{4}{l}{\textit{Filesystem and User}} \\
            AI & Filesystem: Lustre          & AR & Filesystem: Lustre  \\
            AY & User: Permission            &    &                      \\
            \bottomrule
            \end{tabular}
            \label{tab:error_labels}
        \end{table}
        }
        
        \subsubsection{Error Distribution Across Scientific Domains}
\begin{figure}
            \centering
            \includegraphics[width=0.95\linewidth]{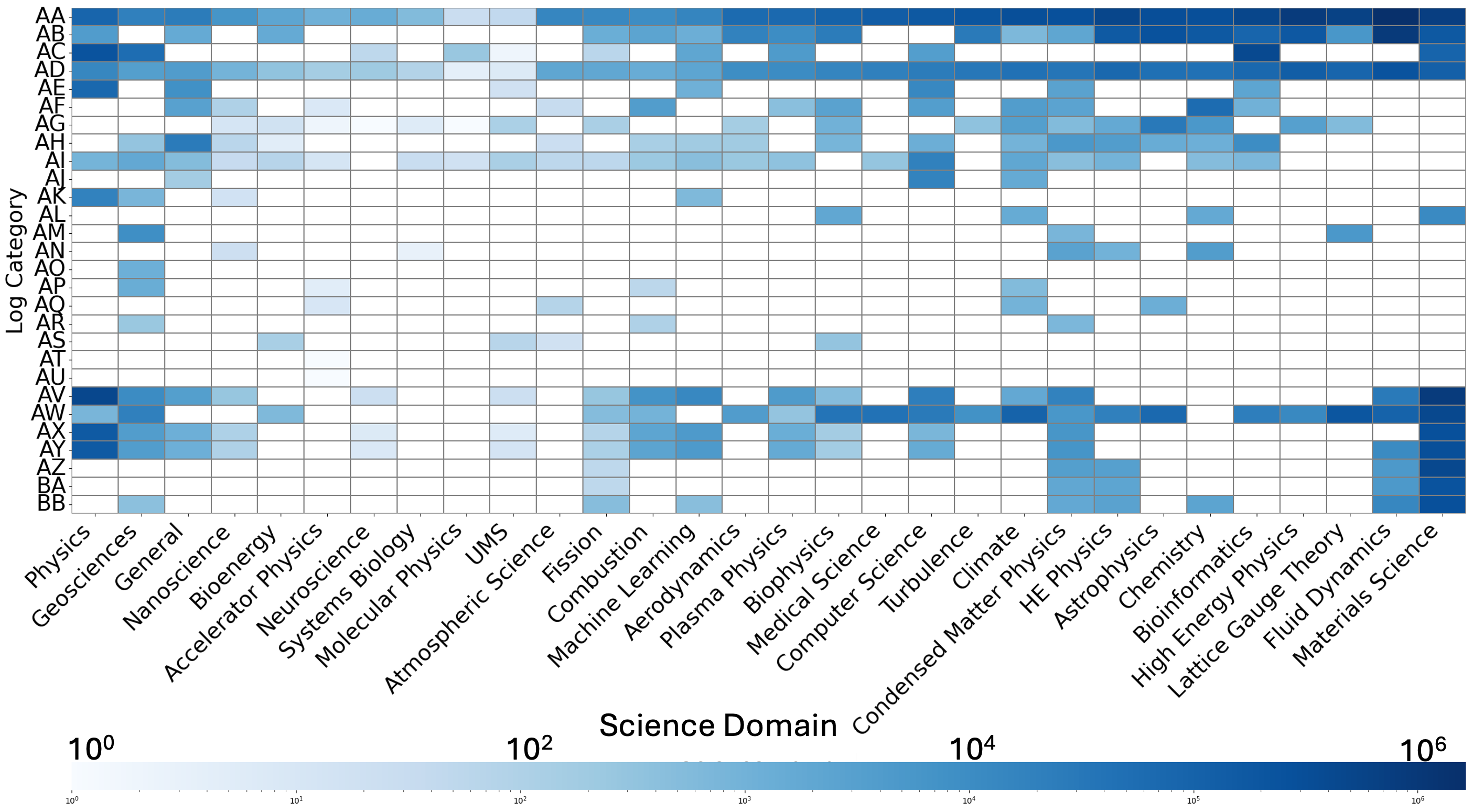}
                    \vspace{-0.1in}

            \caption{28 log categories and 30 science domains are reordered using hierarchical clustering with Ward’s linkage method~\cite{ward1963hierarchical} to identify shared error signatures. Color intensity represents log frequency on a logarithmic scale, highlighting both high-volume system noise and sparse critical faults. Log categories are described in Table~\ref{tab:error_categories}.}
            \label{fig:jobDomain}
        \end{figure}

        Hierarchical clustering shown in Figure~\ref{fig:jobDomain}, using Ward’s method~\cite{ward1963hierarchical} uncovers distinct log signatures that link scientific workloads on the Frontier supercomputer to specific subsystem failure modes. The analysis shows a universal network noise exemplified by AA (Network: Timeout) and AC (Network: Connect Fail). In contrast, specific high-density clusters in the AV to AY range are concentrated in communication and compute heavy workloads suggesting that these domain drive specialized errors such as AX (Network: PCIe) and AV (GPU: General).
        \textit{Our observation is that our methodology provides emperical evidence to explore to domain-aware proactive maintenance, targeting the specific architectural bottlenecks triggered by different scientific workloads.}
        
\subsection{Robustness to Log Noise and Format Evolution}

    To evaluate LLM robustness under realistic log noise and adaptability to evolving systems, we generated transformed variants of Frontier log messages using seven perturbation types simulating common data noise. We selected 100 unique log\_patterns comprising of $3,129$ distinct log messages and applying perturbation, we created $21,903$ test samples. The perturbations spanned three levels: character-level (single-character substitutions like \texttt{kfi\_cxi} $\rightarrow$ \texttt{kfi'cxi}, and parameter changes replacing variables), token-level (missing words through random deletion, via prefix/suffix additions like \texttt{warning:}, and \texttt{word reordering}), and structural (whitespace manipulation through removal/doubling/tab-substitution, and punctuation changes via systematic delimiter substitution such as $() \leftrightarrow {} []$). This approach tests model's ability to handle format variations introduced by software library updates, new data sources, or evolving logging tools.

    Table~\ref{tab:llm_robust} summarizes the robustness results. While we defined accuracy as the exact match rate,  it varies significantly from 96.5\% (parameter changes) to 8.5\% (extra words), the average similarity (Equation~\ref{eq:avg_sim}) remains high (0.844–0.997). 
    \begin{equation}
        \scriptsize
        \text{Avg. Similarity} = \frac{1}{N} \sum_{i=1}^{N} \frac{2 \times M_i}{|p_i^{\text{1}}| + |p_i^{\text{2}}|}
        \label{eq:avg_sim}
    \end{equation}
    where $M$ number of matching characters in longest common subsequence,nd $p^1$ and $p^2$ are two log patterns genereated from original log messages and transformed log messages.
    This dissociation between accuracy and similarity indicates that failures often produce structurally similar but semantically incorrect patterns. 
    The character-level metric (Levenshtein\cite{levenshtein}) and token-level metric (WER\cite{WER}) exhibit distinct behaviors: punctuation changes show minimal corruption (Levenshtein = 0.016, WER = 0.0\%) yet cause substantial accuracy degradation (57.1\%), while whitespace transformations exhibit the highest token-level damage (WER = 10.0\%) but maintain moderate accuracy (78.8\%). 
    Table~\ref{tab:llm_robust} also illustrates specific failure modes through one real example for each case, showing why high similarity does not guarantee correct extraction. The majority of failures exhibit over 98\% similarity, indicating these unmatched patterns differ only in minor details. 
    \textit{We infer that models robustness is driven by semantic impact rather than syntactic extent.}
            \vspace{-0.1in}

\section{Conclusion}

    In this work, we show that instruction-following LLMs, when fine-tuned with a balanced combination of domain-specific log templates and general instruction-following data, provide an accurate and practical solution for HPC log parsing. By jointly leveraging log-template supervision (\textit{SET 1}) and instruction-centric examples (\textit{SET 2}), the combined dataset (\textit{SET 3}) enables a compact 8B-parameter LLM to generalize across heterogeneous and evolving log formats. This approach achieves parsing coverage comparable to that of significantly larger open-source and commercial models. Our evaluation on LogHub benchmarks demonstrates that mixed fine-tuning consistently outperforms using either dataset alone. We show that the fine-tuned smaller models' accuracy performance is similar to large proprietary models with significantly efficient runtime and reduced energy cost by over 20x (Figure~\ref{fig:plot-profiling}). We leverage the finetuned LLM to parse 600 million+ production logs from the Frontier supercomputer, uncovering meaningful operational insights such as temporal burst patterns (Figure~\ref{fig:network_corelation}), failure cascades (Figure~\ref{fig:temporal_corelation_of_events}), workload-specific (Figure~\ref{fig:jobDomain} ), and network's spatial error correlations (Figure~\ref{fig:spatialAnalysis}). By enabling privacy-preserving, locally deployable analysis with compact models, this work establishes instruction-tuned LLMs as a scalable foundation for reliable HPC log parsing and system-level insights such as the multi-stage progression of failure cascades and the interconnect’s critical failure point for leadership-class HPC systems.

\bibliographystyle{ieeetr}
\bibliography{reference}
\end{document}